\documentclass[lettersize,journal]{IEEEtran}
\usepackage{amsmath,amsfonts}
\usepackage{algorithmic}
\usepackage{algorithm}
\usepackage{array}
\usepackage[caption=false,font=normalsize,labelfont=sf,textfont=sf]{subfig}
\usepackage{textcomp}
\usepackage{stfloats}
\usepackage{url}
\usepackage{verbatim}
\usepackage{graphicx}
\usepackage[justification=centering]{caption}
\usepackage{cite}
\begin{document}

% Global variable
\def \w{2.24in}
\def \s{1pt}
\def \etal{\textit{et al.}}
\def \iou{\mathcal{L}_{IoU}}
\def \imean{\overline{\iou}}
\def \r{\mathcal{R}}
\def \grad{\frac{\partial \iou}{\partial W_i}}
\def \wiou#1{\mathcal{L}_{WIoUv{#1}}}
\def \out{\frac{\iou^*}{\ \imean\ }}
\def \ardiff{(\tan^{-1}\frac{w}{h} - \tan^{-1}\frac{w_{gt}}{h_{gt}})}
\def \cenconnect{(x-x_{gt})^2 + (y-y_{gt})^2}
\def \diagsqr{W^2_g + H^2_g}

\title{Wise-IoU: Bounding Box Regression Loss \\ with Dynamic Focusing Mechanism}

\author{Zanjia Tong, Yuhang Chen, Zewei Xu, Rong Yu$^{\ast}$ \thanks{*Correspondingauthor}}
% ,~\IEEEmembership{Staff,~IEEE}

\markboth{}%
{Shell \MakeLowercase{\textit{et al.}}: A Sample Article Using IEEEtran.cls for IEEE Journals}

\IEEEpubid{0000--0000/00\$00.00~\copyright~2021 IEEE}

\maketitle

\begin{abstract}
The loss function for bounding box regression (BBR) is essential to object detection. Its good definition will bring significant performance improvement to the model. Most existing works assume that the examples in the training data are high-quality and focus on strengthening the fitting ability of BBR loss. If we blindly strengthen BBR on low-quality examples, it will jeopardize localization performance. Focal-EIoU v1 was proposed to solve this problem, but due to its static focusing mechanism (FM), the potential of non-monotonic FM was not fully exploited. Based on this idea, we propose an IoU-based loss with a dynamic non-monotonic FM named Wise-IoU (WIoU). The dynamic non-monotonic FM uses the outlier degree instead of IoU to evaluate the quality of anchor boxes and provides a wise gradient gain allocation strategy. This strategy reduces the competitiveness of high-quality anchor boxes while also reducing the harmful gradient generated by low-quality examples. This allows WIoU to focus on ordinary-quality anchor boxes and improve the detector's overall performance. When WIoU is applied to the state-of-the-art real-time detector YOLOv7, the $AP_{75}$ on the MS-COCO dataset is improved from 53.03\% to 54.50\%. Code is available at \textit{\url{https://github.com/Instinct323/wiou}}.
\end{abstract}

\begin{IEEEkeywords}
    Object detection, bounding box regression, dynamic non-monotonic focusing mechanism, generalization performance
\end{IEEEkeywords}

% 1-------------------------------------------------
\section{Introduction}
\IEEEPARstart{T}{he} real-time detectors of the YOLO series have been recognized by most researchers and applied in many scenarios since their inception. Such as YOLOv1 \cite{yolov1}, which constructs a loss function weighted by BBR loss, classification loss, and objectness loss. Until now, this construction is still the most effective loss function paradigm \cite{yolov1, yolov2, yolov3, yolov4, yolov7, fcos1, fcos2, pryolov1} for object detection tasks, where the BBR loss directly determines the localization performance of the model. To further improve the localization performance of the model, a well-designed BBR loss is essential.

% 1.1-----------------------------------------------
\subsection{$\pmb{l_n-}$norm Loss}
For the anchor box $ \vec{B}=[x\ y\ w\ h]$, the values in it correspond to the center coordinates and size of the bounding box. Similarly, $ \vec{B_{gt}}=[x_{gt}\ y_{gt}\ w_{gt}\ h_{gt}]$ describes the properties of the target box. 

YOLOv1 \cite{yolov1} and YOLOv2 \cite{yolov2} are quite similar in the definition of BBR loss. YOLOv2 defines the BBR loss as:
\begin{equation}
    L(\vec B,\vec{B_{gt}}) =\left| \vert  \vec{B} - \vec{B_{gt}} \vert \right|
\end{equation}

However, this form of the loss function cannot shield the interference of the size of the bounding box, making YOLOv2 \cite{yolov2} poor localization performance for small objects. Although YOLOv3 \cite{yolov3} constructs $2-w_{gt} h_{gt}$ in an attempt to reduce the model's attention to large objects, the localization performance brought by this BBR loss to the model is still very limited.

% figure 1
\begin{figure}[t]
    \centering
    \includegraphics[width=\w,scale=1]{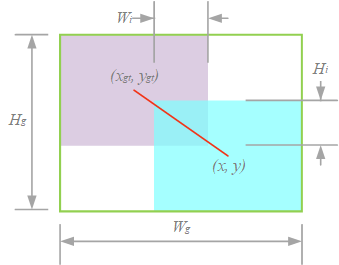}
    \caption{The smallest enclosing box (green) and the central points' connection (red), where the area of the union is $S_u = wh + w_{gt} h_{gt} - W_i H_i$.}
    \label{fig:1-2}
\end{figure}

% 1.2 -------------------------------------------------
\subsection{Intersection over Union}
Intersection over Union \cite{iou} (IoU) is used to measure the degree of overlap between the anchor box and the target box in the object detection task. It effectively shields the interference of bounding box size in the form of proportion, which makes the model can well balance the learning of large objects and small objects when $\iou$ (Eq. \ref{eq:1-2a}) is used as the BBR loss.
\begin{equation}
    \iou  = 1 - IoU = 1 - \frac{W_i H_i}{S_u}
    \label{eq:1-2a}
\end{equation}

However, $\iou$ has another fatal flaw which can be observed in Eq. \ref{eq:1-2b}. $\grad = 0$ when there is no overlap between the bounding boxes ($W_i=0$ or $H_i=0$), the gradient back-propagated by $\iou$ vanishes. As a result, the width of the overlapping region $W_i$ (Fig. \ref{fig:1-2}) cannot be updated during training.
\begin{equation}
    \grad =
    \left\{
        \begin{split}
        -H_i\frac{IoU + 1}{S_u},W_i >0\\
        0,W_i = 0
        \end{split}
    \right.
    \label{eq:1-2b}
\end{equation}

Existing works \cite{giou, diou, eiou, siou} consider many geometric factors related to the bounding box and construct the penalty term $\r_i$ to solve this problem. The existing BBR loss follows the following paradigm:
\begin{equation}
    \mathcal{L}_i = \iou + \r_i
    \label{eq:1-2c}
\end{equation}

% figure 2
\begin{figure*}[!t]
    \centering
    \subfloat[]{
        \includegraphics[width=\w]{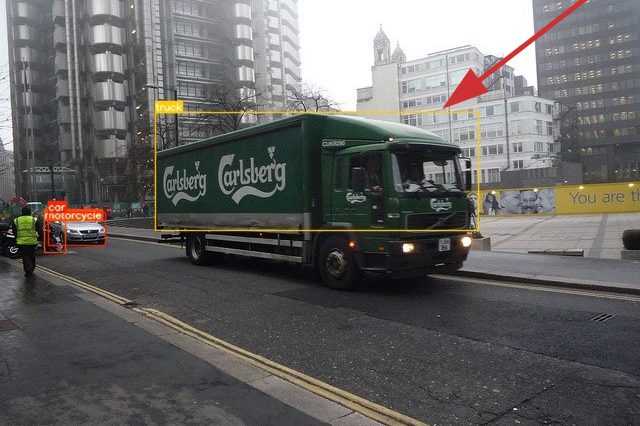}
    }
    \hspace{\s}
    \subfloat[]{
        \includegraphics[width=\w]{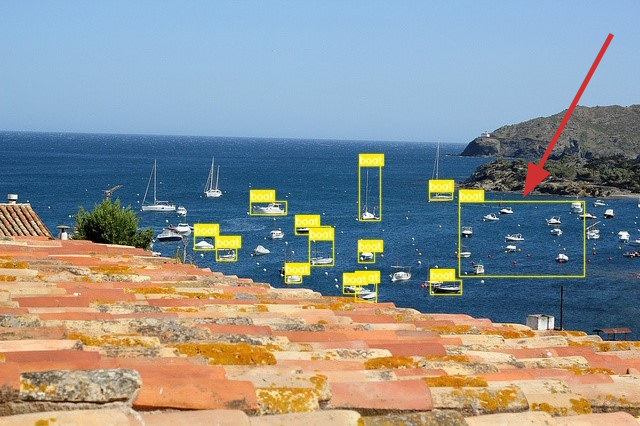}
    }
    \hspace{\s}
    \subfloat[]{
        \includegraphics[width=\w]{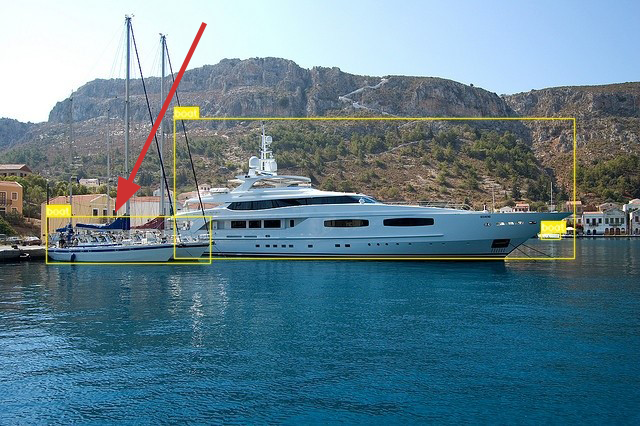}
    }
    \\
    \centering
    \vspace{\s}
    \subfloat[]{
        \includegraphics[width=\w]{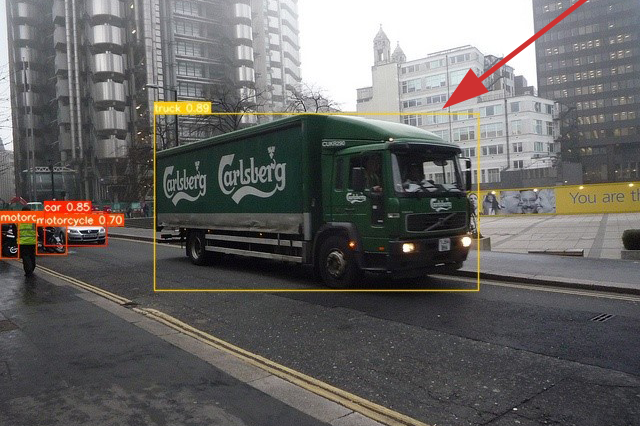}
    }
    \hspace{\s}
    \subfloat[]{
        \includegraphics[width=\w]{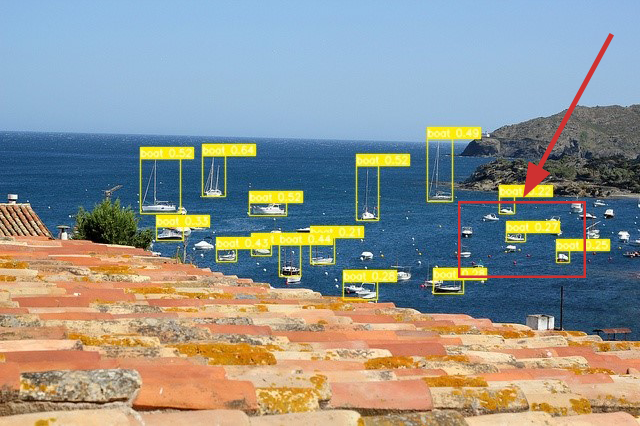}
    }
    \hspace{\s}
    \subfloat[]{
        \includegraphics[width=\w]{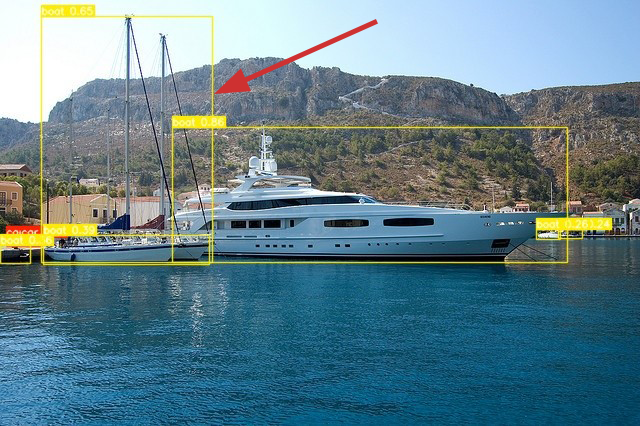}
    }
    \caption{Some target boxes in 1-a, 1-b, and 1-c do not accurately label the instances. Our FM effectively blocks the influence of these low-quality examples and achieves ideal predictions. (a-c) Training data. (d-f) $\wiou{3}$.}
    \label{fig:1-3}
\end{figure*}

\IEEEpubidadjcol
% 1.3 --------------------------------------------------------
\subsection{Focusing Mechanism}
Fig. \ref{fig:1-3} shows some low-quality examples in the training data. A well-performing model will produce large $\iou$ when it produces high-quality anchor boxes for low-quality examples. If the monotonic FM assigns these anchor boxes large gradient gains, the learning of the model will be jeopardized.

In \cite{eiou}, Zhang \etal{} proposed Focal-EIoU v1 using the non-monotonic FM. The FM $f(\iou)$ of Focal-EIoU v1 is static, which specifies the boundary value of the anchor boxes so that the anchor box with $\iou$ equal to the boundary value has the highest gradient gain. Focal-EIoU v1 does not notice that the quality evaluation of anchor boxes is reflected in mutual comparison. It does not fully exploit the potential of the non-monotonic FM.

We define a dynamic FM $f(\beta)$ by estimating the outlier degree of the anchor box as $\beta=\frac{\iou}{\ \imean\ }$. Our FM enables the BBR to focus on ordinary-quality anchor boxes by assigning small gradient gains to high-quality anchor boxes with small $\beta$. At the same time, this mechanism assigns small gradient gains to low-quality anchor boxes with large $\beta$, which effectively weakens the harm of low-quality examples to the BBR.

We combine such a wise FM with IoU-based loss and call it Wise-IoU (WIoU). To evaluate our proposed method, we incorporate WIoU into the state-of-the-art real-time detector YOLOv7 \cite{yolov7}. The main contributions of this paper are summarized as follows:

\begin{itemize}
    \item We propose the attention-based loss WIoU v1 for BBR, which achieves a lower regression error than the state-of-the-art SIoU \cite{siou} in simulation experiments.
    
    \item We design WIoU v2 with monotonic FM, and WIoU v3 with dynamic non-monotonic FM. Benefiting from the wise gradient gain allocation strategy of the dynamic non-monotonic FM, WIoU v3 achieves superior performance.
    
    \item We perform a series of detailed studies of the influence of low-quality examples, demonstrating the effectiveness and efficiency of the dynamic non-monotonic FM.
\end{itemize}

% 2 -------------------------------------------------------------
\section{Related work}
% 2.1 -----------------------------------------------------------
\subsection{Loss Functions for BBR}
To compensate for the scale sensitivity of the $l_2$-norm loss, YOLOv1 \cite{yolov1} weakens the influence of large bounding boxes by performing square root transformation on the size of the bounding boxes. YOLOv3 \cite{yolov3} proposes to construct a penalty term to reduce the competitiveness of large boxes. However, the $l_2$-norm loss ignores the correlation between the bounding box's properties, making this type of BBR loss less effective.

To solve the gradient vanishing problem of IoU loss, GIoU \cite{giou} uses the penalty term constructed by the smallest enclosing box. DIoU \cite{diou} uses the penalty term constructed by distance metric, and CIoU \cite{diou} is obtained by adding the aspect ratio metric based on the DIoU. Gevorgyan constructs SIoU \cite{siou} with angle cost, distance cost, and shape cost, which has a faster convergence rate and better performance.

% 2.2 ------------------------------------------------------
\subsection{Loss Functions with FM}
Cross-entropy loss is widely used in binary classification tasks. However, a salient property of this loss function is that even easy examples produce a large loss value, competing with hard examples. Lin \etal{} proposed focal loss \cite{fl} with the monotonic FM, which effectively reduces the competitiveness of easy examples.

In \cite{eiou}, Zhang \etal{} proposed Focal-EIoU v1 with the non-monotonic FM and Focal-EIoU with the monotonic FM. In their experiments, monotonic FM was shown to be a better choice than non-monotonic FM.

The FM of Focal-EIoU v1 is static, which specifies the quality demarcation standard of the anchor boxes. It gives the highest gradient gain to the anchor box when the IoU loss of the anchor box equals the bound value. It is not noticed that the quality evaluation of the anchor boxes is reflected in the intercomparison, so it does not fully exploit the potential of the non-monotonic FM.

% figure n
\begin{figure*}[t]
    \centering
    \subfloat[]{
        \includegraphics[width=3.0in]{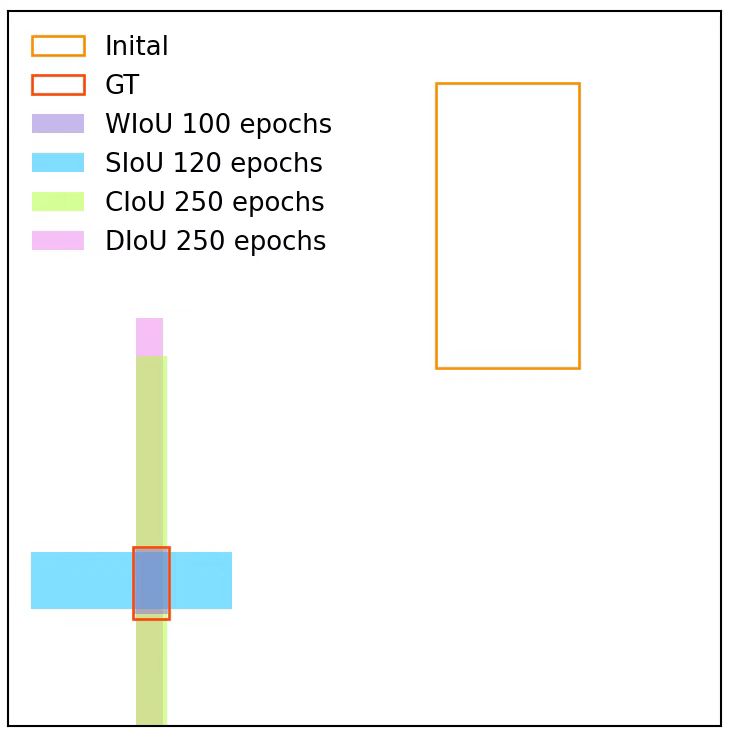}
        \label{fig:3-2a}
    }
    \subfloat[]{
        \includegraphics[width=3.01in]{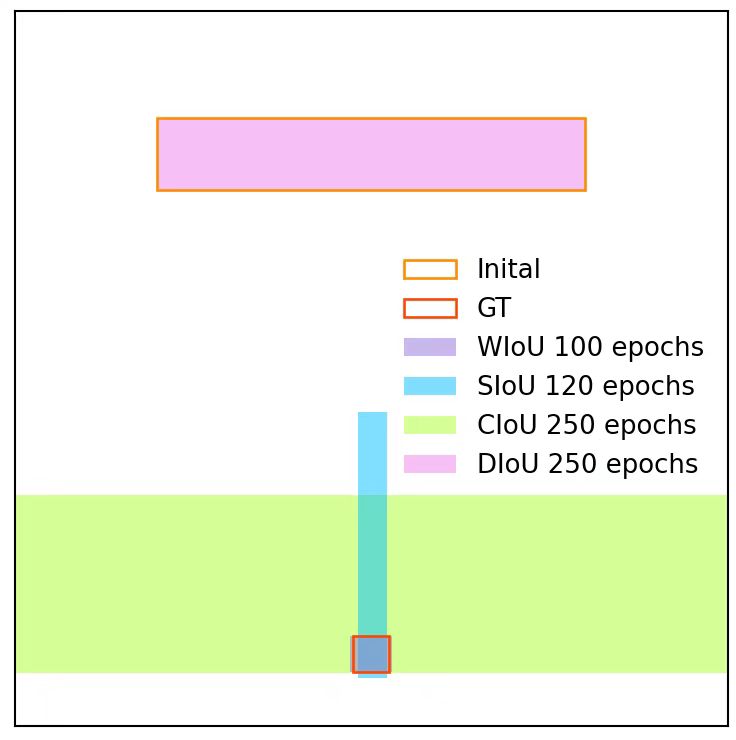}
        \label{fig:3-2b}
    }
    \caption{Regression results with different BBR losses guiding. It is clear that WIoU is optimal.}
    \label{fig:3-2}
\end{figure*}

% 3 -------------------------------------------------------
\section{Method}
% 3.1 -----------------------------------------------------
\subsection{Simulation Experiment}
\label{sec:3.1}
To preliminarily compare each loss function for BBR, we use the simulation experiment proposed by Zheng \etal{} \cite{diou} for evaluation. We generate target boxes (all with area 1/32) at (0.5, 0.5) with 7 aspect ratios (i.e., 1:4, 1:3, 1:2, 1:1, 2:1, 3:1, 4:1). In circular region centered at (0.5, 0.5) with radius $r$, $20000 r^2$ anchor points are uniformly generated. Meanwhile, 49 anchor boxes with 7 scales (i.e., 1/32, 1/24, 3/64, 1/16, 1/12, 3/32, 1/8) and 7 aspect ratios (i.e., 1:4, 1:3, 1:2, 1:1, 2:1, 3:1, 4:1) are placed for each anchor points. Each anchor box needs to be fitted to each target box, and there are 6860000$r^2$ regression cases. To compare the convergence rate in different periods, we set up the following experimental environments: 

% figure 3
\begin{figure}[h]
    \centering
    \subfloat[]{
        \includegraphics[width=1.6in]{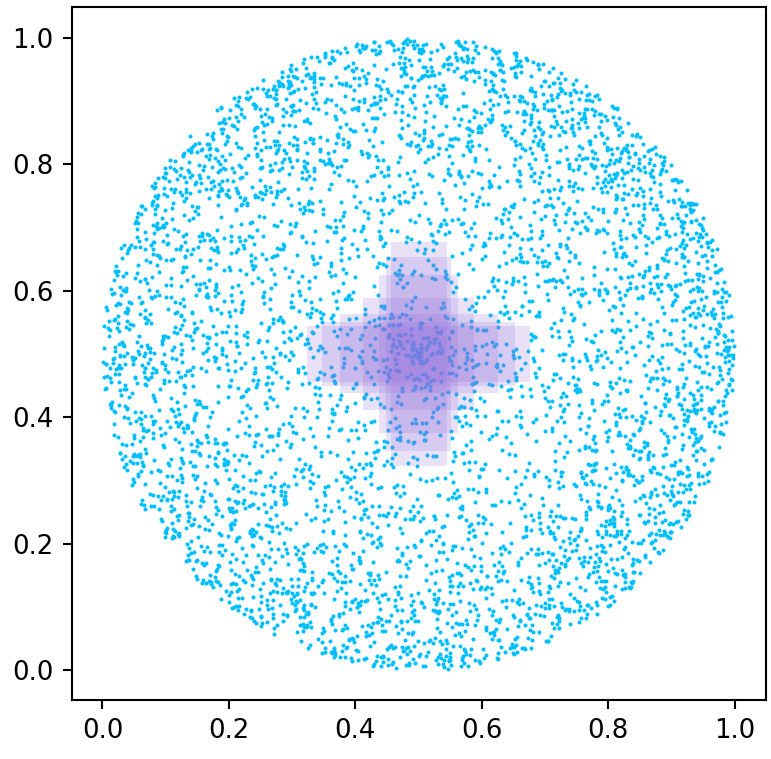}
        \label{fig:3-1a}
    }
    \subfloat[]{
        \includegraphics[width=1.6in]{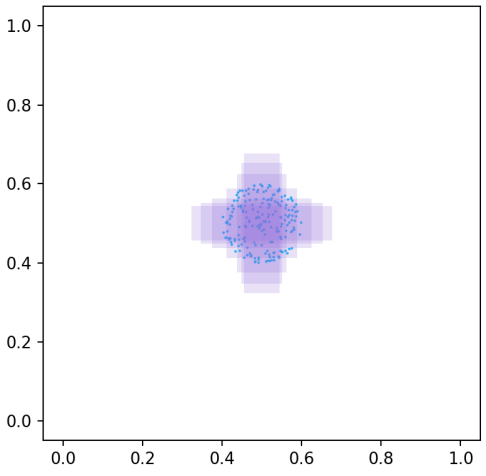}
        \label{fig:3-1b}
    }
    \caption{Anchor points (blue) and target boxes (purple) in simulation experiments. (a) All cases. (b) Major cases. }
    \label{fig:3-1}
\end{figure}

\begin{enumerate}
    \item $r=0.5$, the anchor boxes are distributed inside and outside the coverage area of the target box (Fig. \ref{fig:3-1a}), corresponding to all cases in the BBR.

    \item $r=0.1$, the anchor boxes are generated within the coverage of the target box (Fig. \ref{fig:3-1b}), corresponding to major cases in the BBR.
\end{enumerate}

We also define the loss value as $\overline{\mathcal{L}_i}$ overall regression cases and optimize it by using the gradient descent algorithm with a learning rate of 0.01.

% 3.2-------------------------------------------------------------------------
\subsection{The Solutions of Gradient Vanishing Problem}
Existing BBR losses \cite{giou, diou, eiou, siou} are addition-based and follow the paradigm as shown in Eq. \ref{eq:1-2c}.

% 3.2.1-----------------------------------------------------------------------
\textbf{Distance IoU}: Zheng \etal{} defined $\r_{DIoU}$ \cite{diou} as the normalized distance between central points of two bounding boxes:
\begin{equation}
    \r_{DIoU} = \frac{\cenconnect}{\diagsqr} 
\end{equation}

This term not only solves the gradient vanishing problem of $\iou$, but also serves as a geometric factor. $\r_{DIoU}$ allows DIoU to make a more intuitive choice when faced with anchor boxes with the same $\iou$.
\begin{equation}
    \begin{split}
        \frac {\partial \r_{DIoU}} {\partial W_g} = -2W_g \frac{\cenconnect}{(\diagsqr)^2} < 0 \\
        \frac {\partial \r_{DIoU}} {\partial H_g} = -2H_g \frac{\cenconnect}{(\diagsqr)^2} < 0
    \end{split}
    \label{eq:3-2-1a}
\end{equation}

Meanwhile, $\r_{DIoU}$ provides a negative gradient for the size of the smallest enclosing box, which will make $W_g$ and $H_g$ increase and hinder the overlap between the anchor box and the target box. However, there is no denying that distance metric is indeed an extremely effective solution and becomes a necessary metric for BBR \cite{eiou, siou}. On this basis, Zhang \etal{} increased the punishment for distance metric and proposed EIoU \cite{eiou}:
\begin{equation}
    \r_{EIoU} = \r_{DIoU} + \frac{(x-x_{gt})^2}{W^2_g} + \frac{(y-y_{gt})^2}{H^2_g}
\end{equation}

% 3.2.2-----------------------------------------------------------------------
\textbf{Complete IoU}: On the basis of $\r_{DIoU}$, Zheng \etal{} added the consideration of aspect ratio and proposed $\r_{CIoU}$ \cite{diou}:
\begin{equation}
    \r_{CIoU} = \r_{DIoU} + \alpha v, \alpha = \frac{v}{\iou + v}
\end{equation}
where $v$ describes the consistency of the aspect ratio:
\begin{equation}
    v=\frac{4}{\pi^2} \ardiff ^2
\end{equation}
\begin{equation}
    \begin{split}
        &\frac{\partial v}{\partial w} = \frac{8}{\pi^2} \ardiff ^2 \frac{h}{h^2+w^2}\\
        &\frac{\partial v}{\partial h} = -\frac{8}{\pi^2} \ardiff ^2 \frac{w}{h^2+w^2}
    \end{split}
\end{equation}

Zhang \etal{} \cite{eiou} argued that the irrationality of CIoU is that $\frac{\partial v}{\partial h} = -\frac{w}{h} \frac{\partial v}{\partial w}$,  which means that $v$ cannot provide gradients of the same sign for the width $w$ and height $h$ of the anchor box. In the previous analysis of DIoU, it can be seen that $\r_{DIoU}$ will produce a negative gradient $\frac{\partial \r_{DIoU}}{\partial W_g}$ (Eq. \ref{eq:3-2-1a}). When this negative gradient exactly offsets the gradient generated by $\iou$ on the anchor box, the anchor box will not be optimized. The consideration of aspect ratio by CIoU will break this deadlock (Fig. \ref{fig:3-2b}).

% 3.2.3-----------------------------------------------------------------------
\textbf{Scylla IoU}: Gevorgyan \cite{siou} proved that the center-aligned anchor box would have a faster convergence speed, and constructed SIoU in terms of angle cost, distance cost, and shape cost.

The angle cost describes the minimum angle between the central points' connection (Fig. \ref{fig:1-2}) and the x-y axis:
\begin{equation}
    \mathcal{\varLambda} = \sin(2\sin^{-1}\frac{\min(|x-x_{gt}|,|y-y_{gt}|)}{\sqrt{\cenconnect} + \epsilon})
\end{equation}

When the central points are aligned on the x-axis or y-axis, $\varLambda = 0$. When the central points' connection is at 45° to the x-axis, $\varLambda = 1$. This term can guide the anchor box to drift to the nearest axis of the target box, reducing the total number of degrees of freedom of BBR.

The distance cost describes the distance between central points, and its penalty is positively correlated with the angle cost. The distance cost is defined as:
\begin{equation}
    \Delta = \frac{1}{2}\sum_{t=w,h}(1 - e^{-\gamma \rho_{t}}),\gamma = 2 - \varLambda 
\end{equation}
\begin{equation}
    \left\{
        \begin{split}
            \rho_x = (\frac{x - x_{gt}}{W_g})^2\\
            \rho_y = (\frac{y - y_{gt}}{H_g})^2
        \end{split}
    \right.
\end{equation}

The shape cost describes the size difference between the bounding boxes. When the sizes of the bounding boxes are inconsistent, $\Omega \neq 0$. And it is defined as:
\begin{equation}
    \Omega = \frac{1}{2}\sum_{t=w,h}(1 - e^{\omega_t})^{\theta},\theta = 4
\end{equation}
\begin{equation}
    \left\{
        \begin{split}
        \omega_w = \frac{|w-w_{gt}|}{\max(w,w_{gt})}\\
        \omega_h = \frac{|h-h_{gt}|}{\max(h,h_{gt})}
        \end{split}
    \right.
\end{equation}

$\r_{SIoU}$ is similar to $\r_{CIoU}$, they both consist of distance cost and shape cost:
\begin{equation}
    \r_{SIoU} = \Delta + \Omega
\end{equation}

Since the penalty of $\r_{SIoU}$ on the distance metric increases with the increase of shape cost, the models trained by SIoU have faster convergence speed and lower regression error.

% 3.3 ----------------------------------------------------------
\subsection{The Proposed Methods}
% 3.3.1-----------------------------------------------------------------------
Because training data inevitably contain low-quality examples, geometric factors such as distance and aspect ratio will aggravate the penalty for low-quality examples and thus degrade the generalization performance of the model. A good loss function should weaken the penalty of geometric factors when the anchor box coincides well with the target box, and less intervention in training will make the model get better generalization ability. Based on this, we construct distance attention (Eq. \ref{eq:3-3a}) and obtain WIoU v1 with two layers of attention mechanism:

\begin{itemize}
    \item $\r_{WIoU} \in [1,e)$, which will significantly amplify $\iou$ of the ordinary-quality anchor box.
    \item $\iou \in [0,1]$, which will significantly reduce $\r_{WIoU}$ of the high-quality anchor box and its focus on the distance between central points when the anchor box coincides well with the target box.
\end{itemize}
\begin{equation}
    \begin{split}
        & \wiou{1} = \r_{WIoU} \iou \\
        & \r_{WIoU} = \exp(\frac{\cenconnect}{(\diagsqr)^*})
    \end{split}
    \label{eq:3-3a}
\end{equation}
where $W_g, H_g$ are the size of the smallest enclosing box (Fig. \ref{fig:1-2}). In order to prevent $\r_{WIoU}$ from producing the gradient that hinders convergence, $W_g, H_g$ are detached from the computational graph (the superscript $^*$ indicates this operation). Because it effectively eliminates the factor that hinders convergence, we do not introduce new metrics such as aspect ratio.

% figure 6
\begin{figure*}[!t]
    \centering
    \subfloat[]{
        \includegraphics[width=\w]{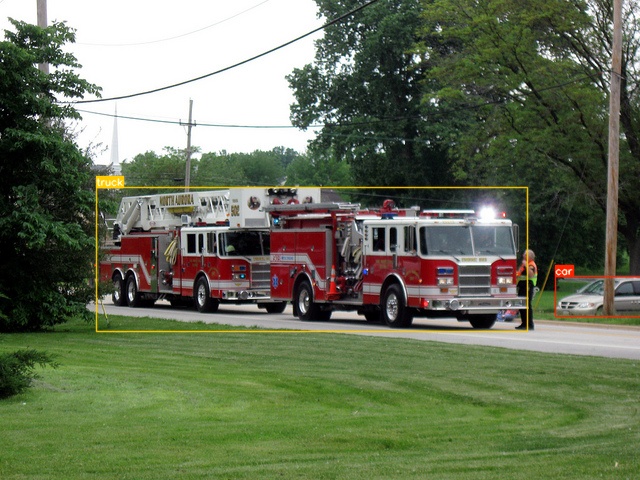}
    }
    \hspace{\s}
    \subfloat[]{
        \includegraphics[width=\w]{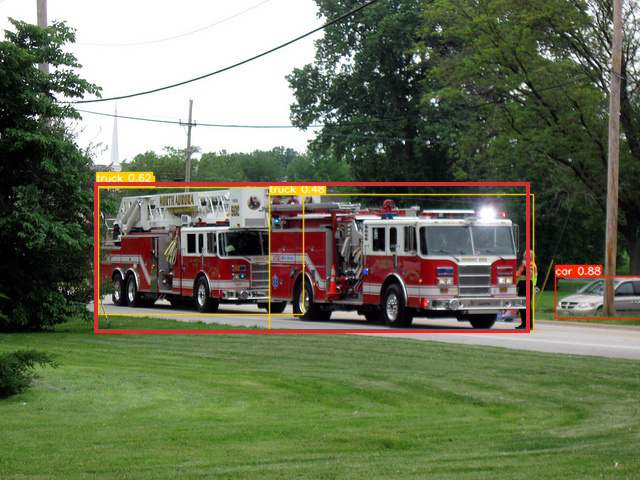}
    }
    \hspace{\s}
    \subfloat[]{
        \includegraphics[width=\w]{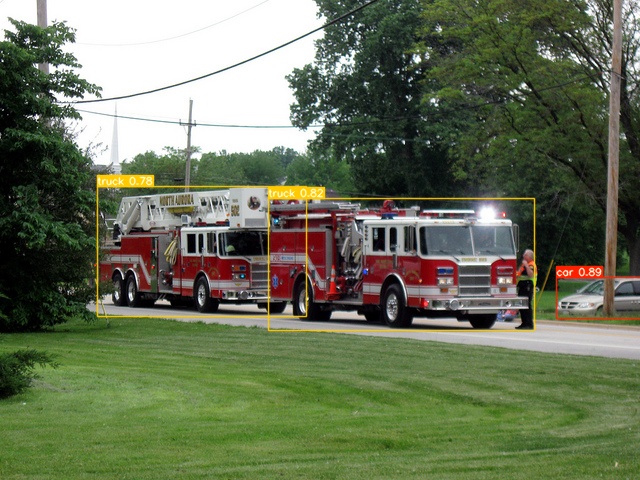}
    }
    \\
    \centering
    \vspace{\s}
    \subfloat[]{
        \includegraphics[width=\w]{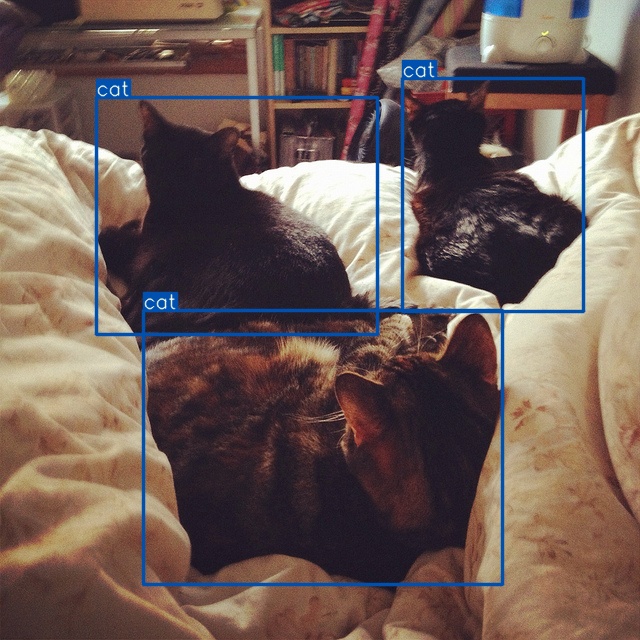}
    }
    \hspace{\s}
    \subfloat[]{
        \includegraphics[width=\w]{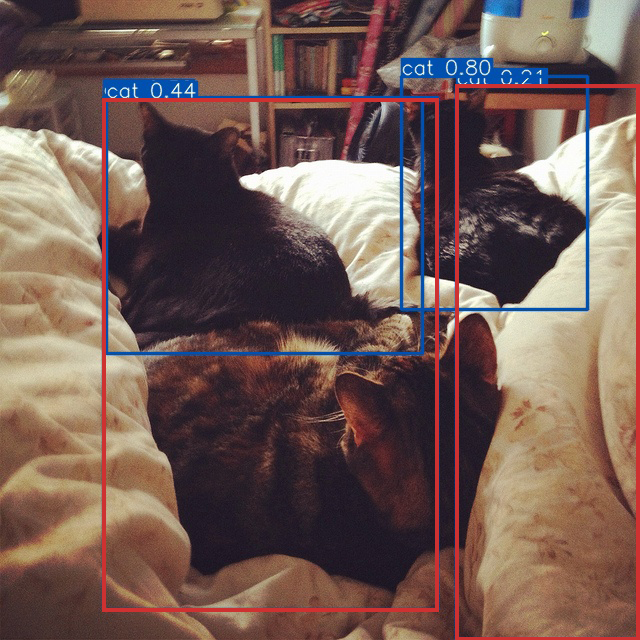}
    }
    \hspace{\s}
    \subfloat[]{
        \includegraphics[width=\w]{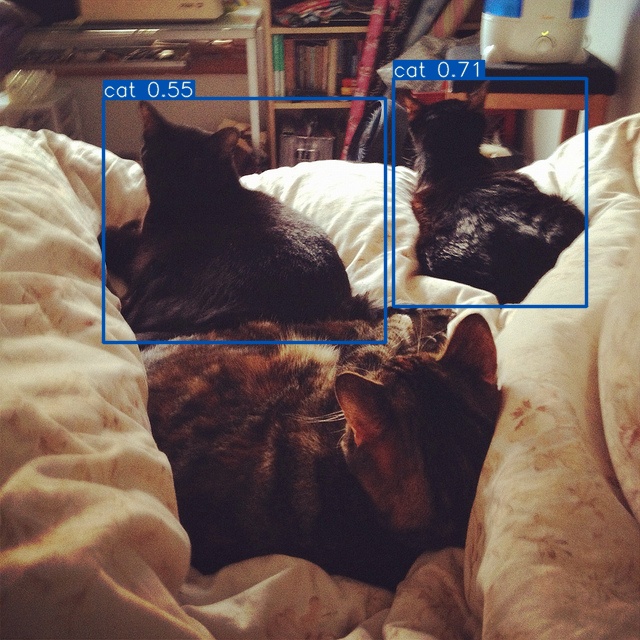}
    }
    \caption{Training results of BBR losses with different FMs. (a,d) Training data. (b,e) $\wiou{2}$. (c,f) $\wiou{3}$.}
    \label{fig:4-2}
\end{figure*}

\begin{figure}[h]
    \centering
    \hspace{-17pt}
    \subfloat[]{
        \includegraphics[width=1.8in]{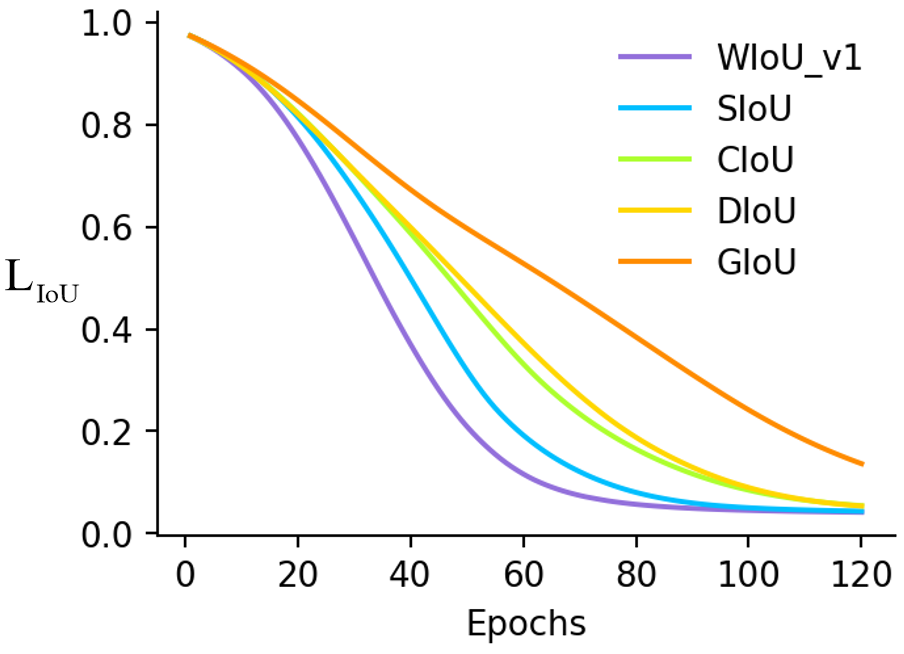}
        \label{fig:3-3a}
    }
    \hspace{-12pt}
    \subfloat[]{
        \includegraphics[width=1.8in]{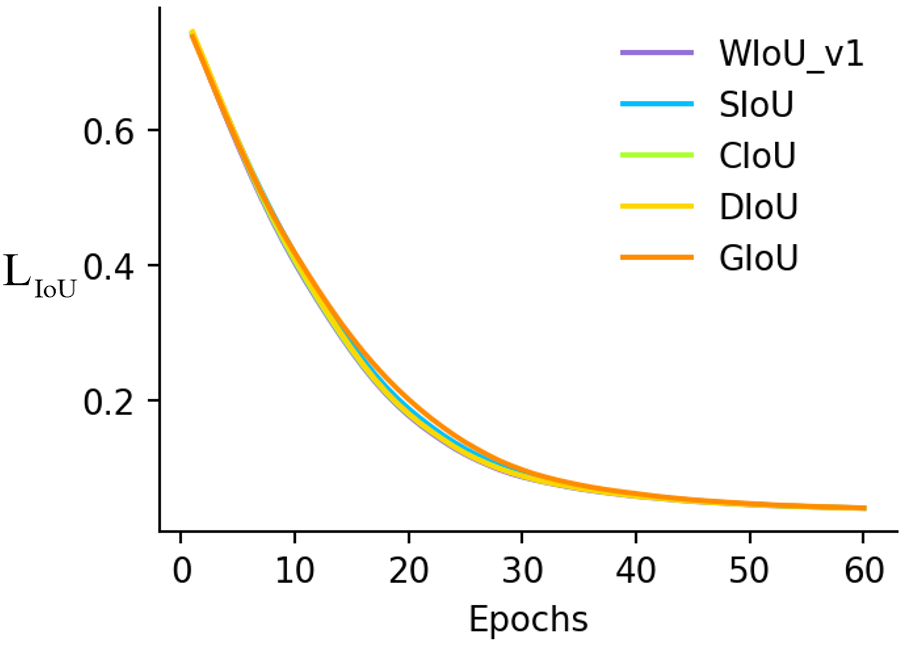}
        \label{fig:3-3b}
    }
    \caption{$\iou$ curves of BBR losses without FMs in simulation experiments. (a) All cases. (b) Major cases.}
    \label{fig:3-3}
\end{figure}

Through the simulation experiment mentioned in \ref{sec:3.1}, we compare the performance of BBR losses without FMs. From the results of Fig. \ref{fig:3-3}, we have the following observations:

\begin{enumerate}
    \item Among a series of BBR losses mentioned in existing works, SIoU \cite{siou} has the fastest convergence rate.
    
    \item For the main cases in the BBR, all BBR losses have extremely similar convergence rates. It follows that the difference in convergence rate mainly comes from non-overlapping bounding boxes. Our proposed attention-based WIoU v1 has the best effect on this part.
\end{enumerate}

% figure 4

% 3.3.2-----------------------------------------------------------------------
\textbf{Learning from focal loss}: Lin \etal{} \cite{fl} designs a monotonic FM for cross-entropy, which effectively reduces the contribution of easy examples to the loss value. Thus, the model can focus on hard examples and obtain classification performance improvement. Similarly, we construct monotonic focusing coefficient $\iou^{\gamma *}$ for $\wiou{1}$.
\begin{equation}
    \wiou{2} = \iou^{\gamma *} \wiou{1}, \gamma > 0
\end{equation}

Due to the addition of the focusing coefficient, the gradient back-propagated by WIoU v2 also changes:
\begin{equation}
    \frac{\partial \wiou{2}} {\partial \iou} = \iou^{\gamma *} \frac{\partial \wiou{1}}{\partial \iou}, \gamma > 0
\end{equation}

Note that the gradient gain is $r = \iou^{\gamma *} \in [0,1]$. During the model's training, the gradient gain decreases with the decrease of $\iou$, resulting in a slow convergence rate in the late stages of training. Therefore, the mean of $\iou$ is introduced as the normalizing factor:
\begin{equation}
    \wiou{2} = (\out)^{\gamma} \wiou{1}
\end{equation}
where $\imean$ is the exponential running average with momentum $m$. Dynamically updating the normalizing factor keeps the gradient gain $r = (\out)^{\gamma}$ at a high level overall, which solves the problem of slow convergence in the late stages of training.

% 3.3.3-----------------------------------------------------------------------
\textbf{Dynamic non-monotonic FM}: The outlier degree of the anchor box is characterized by the ratio of $\iou$ to $\imean$:
\begin{equation}
    \beta = \out \in [0, +\infty)
\end{equation}

\begin{figure}[h]
    \centering
    \hspace{-16pt}
    \includegraphics[width=3.0in,scale=1]{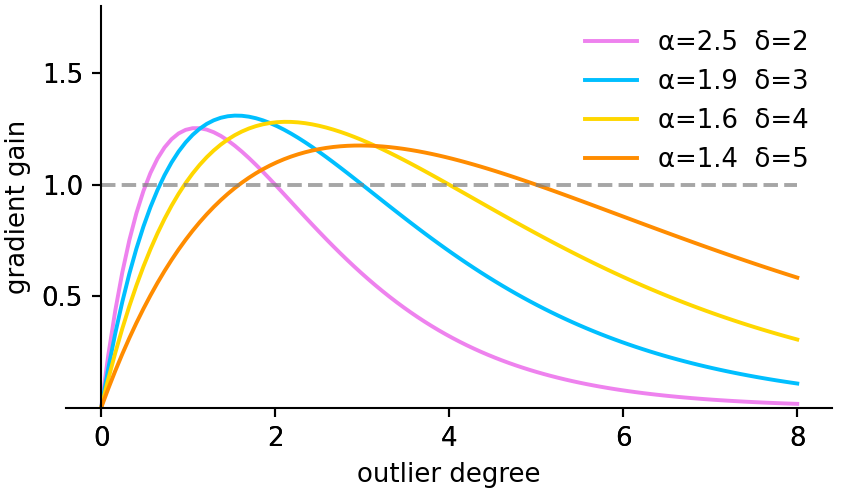}
    \caption{The mapping of outlier degree $\beta$ and gradient gain $r$, which is controlled by the hyper-parameters $\alpha, \delta$.}
    \label{fig:3-3o}
\end{figure}

A small outlier degree means that the anchor box is high-quality. We assign a small gradient gain to it in order to focus the BBR on ordinary-quality anchor boxes. Additionally, assigning a small gradient gain to the anchor box with a large outlier degree will effectively prevent large harmful gradients from low-quality examples. We construct a non-monotonic focusing coefficient using $\beta$ and apply it to WIoU v1:
\begin{equation}
    \wiou{3} = r\wiou{1},\ r=\frac{\beta}{\delta \alpha^{\beta - \delta}}
\end{equation}
where $\delta$ makes $r=1$ when $\beta = \delta$. As shown in Fig. \ref{fig:3-3o}, the anchor box will enjoy the highest gradient gain when its outlier degree satisfies $\beta = C$ ($C$ is a constant value). Since $\imean$ is dynamic, the quality demarcation standard of anchor boxes is also dynamic, which allows WIoU v3 to make the gradient gain allocation strategy that is most in line with the current situation at every moment.

% figure 8(new)
\begin{figure*}[t]
    \centering
    \subfloat[]{
        \includegraphics[width=\w]{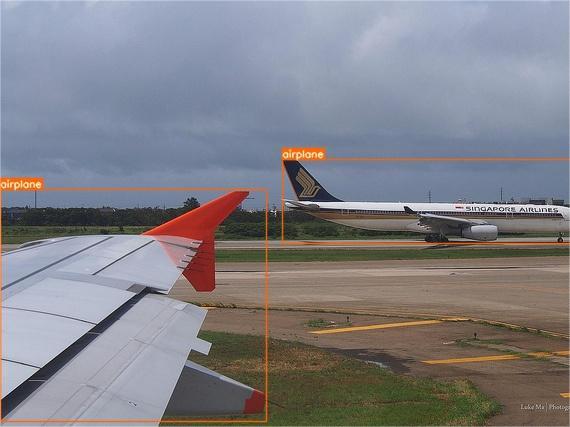}
    }
    \hspace{\s}
    \subfloat[]{
        \includegraphics[width=\w]{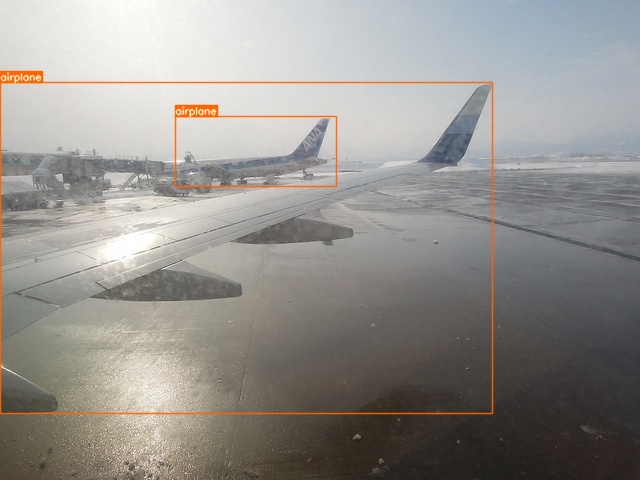}
    }
    \hspace{\s}
    \subfloat[]{
        \includegraphics[width=\w]{4-3c}
    }
    \\
    \centering
    \vspace{\s}
    \subfloat[]{
        \includegraphics[width=\w]{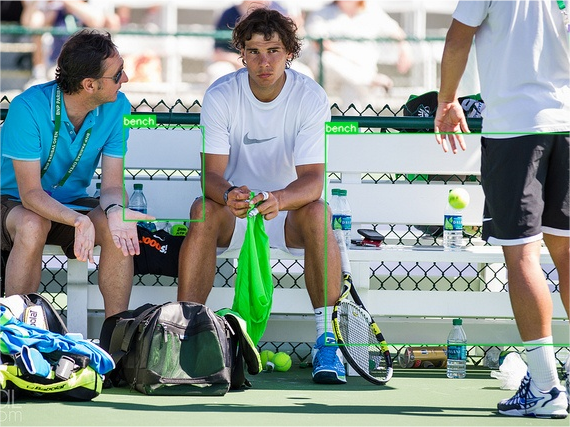}
    }
    \hspace{\s}
    \subfloat[]{
        \includegraphics[width=\w]{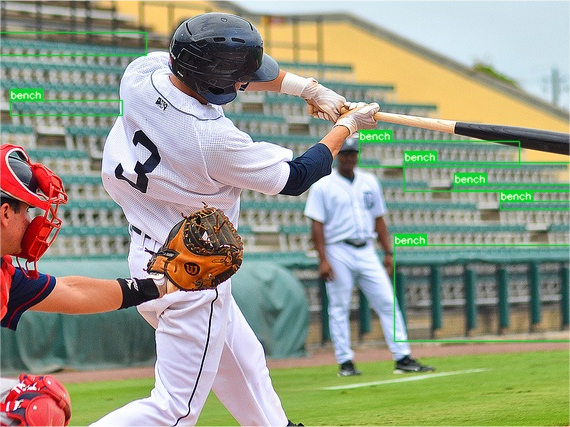}
    }
    \hspace{\s}
    \subfloat[]{
        \includegraphics[width=\w]{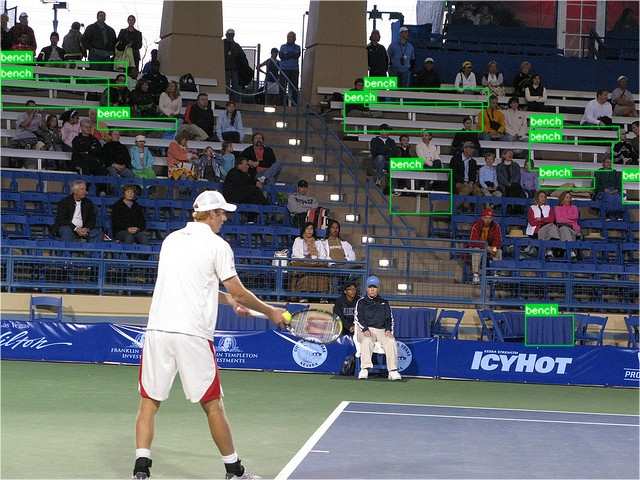}
    }
    \caption{The partial labels of the dataset. (a-c) Controversial label. (d) Error label. (e-f) Incomplete label.}
    \label{fig:4-3}
\end{figure*}

% figure 5

% figure 7
\begin{figure}[t]
    \centering
    \hspace{-20pt}
    \includegraphics[width=1.4in]{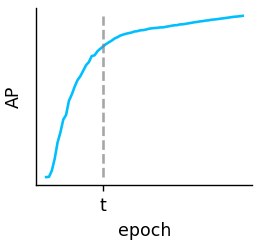}
    \caption{The epoch $t$ when AP's ascension slowed significantly.}
    \label{fig:3-3c}
\end{figure}

To prevent low-quality anchor boxes from being left behind in the early stages of training, we initialize $\imean=1$, making the anchor box that $\iou=1$ enjoys the highest gradient gain. To maintain such a strategy in the early stages of training, it is necessary to set a small momentum $m$ to delay the time when $\imean$ approaches the real value $\overline{\mathcal{L}_{IoU-real}}$. For the training where the number of batches is $n$ and the lifting speed of AP slows significantly at the epoch of $t$ (Fig. \ref{fig:3-3c}), we recommend setting the momentum as:
\begin{equation}
    m = 1-\sqrt[tn]{0.05}
\end{equation}
this setup makes $\imean \approx \overline{\mathcal{L}_{IoU-real}}$ after $t$ epochs of training.

In the middle and late stages of training, WIoU v3 assigns small gradient gains to low-quality anchor boxes to reduce the harmful gradients. At the same time, it also focuses on ordinary-quality anchor boxes to improve the localization performance of the model.

% 4 -----------------------------------------------------------
\section{Experiments}
% 4.1 -------------------------------------------------------------
\subsection{Experimental Setup}
For a fair comparison, all of our experiments are performed on the PyTorch framework \cite{torch}. For the dataset, we select 20 categories in the MS-COCO dataset \cite{coco}, and select 28474 images as the training data and 1219 images as the validation data. For the model, we choose YOLOv7-w6 \cite{yolov7} with the layer channel multiple of 0.75 for training. The models are trained for 120 epochs with batch size 32 and different BBR losses. At the same time, the momentum of $\imean$ is set according to $n = 890, t = 34$.

The anchor boxes produced by the detection heads of YOLOv7 mainly contain two parts: the anchor boxes from lead heads (ABLH) and the anchor boxes from auxiliary heads (ABAH). The ABLH tends to have better-fitting results and less information, and the ABAH is the opposite. If we only count the mean of the ABLH, it will cause the gradient gains of the ABAH to disappear gradually, making the FM ignore the rich information amount of the ABAH. Therefore, our mean statistics include the ABLH and the ABAH.

% figure 7
\begin{figure*}[!t]
    \centering
    \hspace{-20pt}
    \includegraphics[width=7.0in]{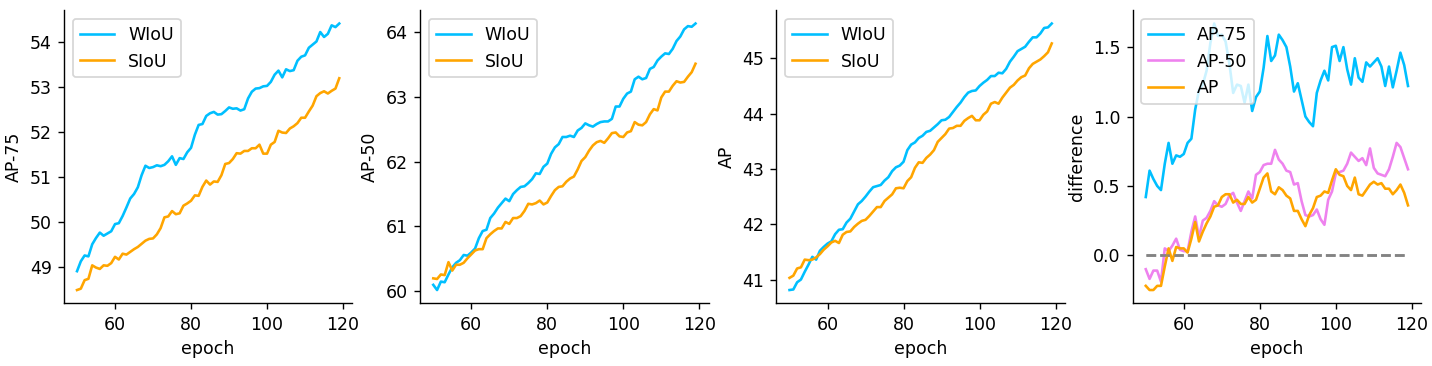}
    \caption{AP curves of the models during training, where the difference of AP is the leading value of the model trained by WIoU v3.}
    \label{fig:4-2o}
\end{figure*}

% table 1
\begin{table*}[t]
    \centering
    \caption{Performance of each bounding box loss.}
    \setlength{\tabcolsep}{3mm}{
    \begin{tabular}{l |  l l l}
        \hline
         & $AP_{75}^{val}$ & $AP_{50}^{val}$  & $AP^{val}$ \\ 
        \hline
        CIoU \cite{diou} & 53.03 & 63.14 & 45.20 \\
        CIoU v2 ($\gamma = 0.5$) & 53.47 (+0.44) & 63.41 (+0.27) & 45.12 \\
        CIoU v3 ($\alpha = 1.4,\delta = 5$) & 53.25 (+0.22) & 63.34 (+0.20) & 44.76 \\
        CIoU v3 ($\alpha = 1.6,\delta = 4$) & 53.68 (+0.65) & 63.34 (+0.20) & 45.10\\
        CIoU v3 ($\alpha = 1.9,\delta = 3$) & 53.04 & 62.92 & 44.91\\
        \hline
        SIoU \cite{siou} & 53.15 & 63.46 & 45.21 \\
        SIoU v2 ($\gamma = 0.5$) & 53.07 & 63.12 & 44.66 \\
        SIoU v3 ($\alpha = 1.4,\delta = 5$) & 53.27 (+0.12) & 64.13 (+0.67) & 45.15 \\
        SIoU v3 ($\alpha = 1.6,\delta = 4$) & 53.21 & 63.48 & 44.89\\
        SIoU v3 ($\alpha = 1.9,\delta = 3$) & 53.42 (+0.27) & 63.28 & 45.03 \\
        \hline 
        EIoU \cite{eiou} & 53.55 & 63.17 & 45.39 \\
        Focal-EIoU \cite{eiou} & 52.88 & 63.37 (+0.20) & 44.75 \\
        \hline 
        WIoU v1 & 52.82 & 63.15 & 44.87 \\
        WIoU v2 ($\gamma = 0.5$) & 53.67 (+0.85) & 64.15 (+1.00) & 45.56 (+0.68) \\
        WIoU v3 ($\alpha = 1.4,\delta = 5$) & 53.75 (+1.07) & 64.05 (+0.90) & 45.15 (+0.28)\\
        WIoU v3 ($\alpha = 1.6,\delta = 4$) & 53.91 (+1.09) & 64.16 (+1.01) & 45.44 (+0.57)\\
        WIoU v3 ($\alpha = 1.9,\delta = 3$) & \textbf{54.50 (+1.68)} & \textbf{64.20 (+1.05)} & \textbf{45.68 (+0.81)}\\
        \hline
    \end{tabular}}
    \label{tab:1}
\end{table*}

\begin{table*}[t]
    \centering
    \caption{The precision of the object detection model for partial categories.}
    \begin{tabular}{c |c c c c c c c c }
        \hline
          & airplane & train & truck & traffic light & stop sign & parking meter & bench & elephant \\
          \hline 
          SIoU & \textbf{74.74} & 78.85 & 51.37 & 54.96 & 58.33 & 58.18 & \textbf{40.27} & 86.32 \\
          Focal-EIoU & 72.92 & \textbf{84.26} & 53.37 & 51.95 & 55.10 & 54.55 & 38.13 & 85.17 \\
          WIoU v3 & 71.58 & 81.82 & \textbf{55.94} & \textbf{55.45} & \textbf{64.58} & \textbf{62.26} &  36.81 & \textbf{87.83} \\
          \hline
    \end{tabular}
    \label{tab:2}
\end{table*}

% 4.2 ----------------------------------------------------------------
\subsection{Ablation Study}
We apply the FMs to BBR losses to investigate the effect of the FMs on addition-based losses. Version 2 of these BBR losses used a setting of $\gamma = 0.5$, to align with the monotonic FM of Focal-EIoU \cite{eiou}. Their version 3 uses the dynamic non-monotonic FM proposed in this paper.

By comparing BBR losses' version 2 with the original version (TABLE \ref{tab:1}), it is known that the monotonic FM negatively affects the performance of both SIoU \cite{siou} and EIoU \cite{eiou}. Because these two punish the distance metric more strongly, larger harmful gradients are synthesized under the action of the monotonic FM. CIoU \cite{diou} and WIoU v1 are less penalized for the distance metric, which allows them to effectively weaken the amplification of the harmful gradient by the monotonic FM.

By comparing BBR losses' version 3 with the original version (TABLE \ref{tab:1}), we can know that the non-monotonic FM can effectively improve the performance of BBR losses. For each BBR loss, there is a set of unique parameters that can maximize this performance gain.

In addition, we compare the regression results for the anchor boxes (Fig. \ref{fig:4-2}). WIoU v2 with monotonic FM is affected by low-quality examples, resulting in poor predictions. WIoU v3 benefits from the dynamic non-monotonic FM, which effectively shields the influence from low-quality examples and achieves ideal predictions.

% 4.3 ----------------------------------------------------------------
\subsection{Comparison Study}
In TABLE \ref{tab:1}, the performance ranking of the BBR losses' original version is: EIoU $>$ SIoU $>$ CIoU $>$ WIoU v1. Such an order also agrees with the strength of their penalties for distance metric. However, when the FMs are applied, the performance gains of BBR losses are in the opposite order. In the experiments, the model trained by WIoU v3 achieved the best performance.

We monitor the change in the precision of YOLOv7 during training (Fig. \ref{fig:4-2o}). Due to the dynamic non-monotonic FM, our proposed WIoU v3 effectively shields many negative effects during training, so the model's precision can increase faster.

After comparing WIoU v3 with the state-of-the-art BBR losses, several categories with large differences in precision are obtained (TABLE \ref{tab:2}). Benefiting from the ability to identify low-quality examples, the model trained by WIoU v3 has greatly improved precision for some categories. At the same time, the model's precision for airplanes and benches has decreased.

Some of the labels of airplanes are controversial (Fig. \ref{fig:4-3}), and some of the selected airplanes lack prominent features such as fuselage. These examples are as hard to learn as low-quality examples, and this part of hard examples is discarded by the FM of WIoU v3. In addition, there are many errors in the labels of benches, and there are also a large number of benches that have not been labeled. This is unfair for models that generalize well and detect more benches. 

% 5 -------------------------------------------------------------------
\section{Conclusion}
In this paper, we observe that the low-quality examples in the training data will hinder the generalization of the object detection model. Most existing works are limited to static focusing mechanism (FM), which does not fully exploit the potential of non-monotonic FM. Although the monotonic FM they advocate can improve localization performance, it does not solve this problem. We propose a dynamic non-monotonic FM that can reduce the competitiveness of high-quality anchor boxes and mask the influence of low-quality examples. 

In the ablation study, we show that dynamic non-monotonic FM leads to better generalization performance for the model. Because WIoU v1 has the adjustment of the penalty term by the attention mechanism, the interaction between WIoU V1 and the dynamic non-monotonic FM can make the model achieve significant performance improvement.

In the comparison study, the model trained by WIoU v3 achieved significant improvement in the precision for some categories. At the same time, the precision for some categories is also reduced due to the low-quality data annotation.

Learning appropriate knowledge with limited parameters is the key to the success of real-time detectors. WIoU v3 improves the overall performance of the detector by weighing the learning of low-quality examples and high-quality examples.


\begin{thebibliography}{1}
\bibliographystyle{IEEEtran}

\bibitem{yolov1}
Joseph Redmon, Santosh Divvala, Ross Girshick, and Ali Farhadi,
\newblock ``You only look once: Unified, real-time object detection,''
\newblock in {\em Proceedings of the IEEE conference on computer vision and
  pattern recognition}, 2016, pp. 779--788.

\bibitem{yolov2}
Joseph Redmon and Ali Farhadi,
\newblock ``Yolo9000: better, faster, stronger,''
\newblock in {\em Proceedings of the IEEE conference on computer vision and
  pattern recognition}, 2017, pp. 7263--7271.

\bibitem{yolov3}
Joseph Redmon and Ali Farhadi,
\newblock ``Yolov3: An incremental improvement,''
\newblock {\em arXiv preprint arXiv:1804.02767}, 2018.

\bibitem{yolov4}
Alexey Bochkovskiy, Chien-Yao Wang, and Hong-Yuan~Mark Liao,
\newblock ``Yolov4: Optimal speed and accuracy of object detection,''
\newblock {\em arXiv preprint arXiv:2004.10934}, 2020.

\bibitem{yolov7}
Chien-Yao Wang, Alexey Bochkovskiy, and Hong-Yuan~Mark Liao,
\newblock ``Yolov7: Trainable bag-of-freebies sets new state-of-the-art for
  real-time object detectors,''
\newblock {\em arXiv preprint arXiv:2207.02696}, 2022.

\bibitem{fcos1}
Zhi Tian, Chunhua Shen, Hao Chen, and Tong He,
\newblock ``Fcos: Fully convolutional one-stage object detection,''
\newblock in {\em Proceedings of the IEEE/CVF international conference on
  computer vision}, 2019, pp. 9627--9636.

\bibitem{fcos2}
Zhi Tian, Chunhua Shen, Hao Chen, and Tong He,
\newblock ``Fcos: A simple and strong anchor-free object detector,''
\newblock {\em IEEE Transactions on Pattern Analysis and Machine Intelligence},
  2020.

\bibitem{pryolov1}
Xiang Long, Kaipeng Deng, Guanzhong Wang, Yang Zhang, Qingqing Dang, Yuan Gao,
  Hui Shen, Jianguo Ren, Shumin Han, Errui Ding, et al.,
\newblock ``Pp-yolo: An effective and efficient implementation of object
  detector,''
\newblock {\em arXiv preprint arXiv:2007.12099}, 2020.

\bibitem{iou}
Jiahui Yu, Yuning Jiang, Zhangyang Wang, Zhimin Cao, and Thomas Huang,
\newblock ``Unitbox: An advanced object detection network,''
\newblock in {\em Proceedings of the 24th ACM international conference on
  Multimedia}, 2016, pp. 516--520.

\bibitem{giou}
Hamid Rezatofighi, Nathan Tsoi, JunYoung Gwak, Amir Sadeghian, Ian Reid, and
  Silvio Savarese,
\newblock ``Generalized intersection over union: A metric and a loss for
  bounding box regression,''
\newblock in {\em Proceedings of the IEEE/CVF conference on computer vision and
  pattern recognition}, 2019, pp. 658--666.

\bibitem{diou}
Zhaohui Zheng, Ping Wang, Wei Liu, Jinze Li, Rongguang Ye, and Dongwei Ren,
\newblock ``Distance-iou loss: Faster and better learning for bounding box
  regression,''
\newblock in {\em Proceedings of the AAAI conference on artificial
  intelligence}, 2020, vol.~34, pp. 12993--13000.

\bibitem{eiou}
Yi-Fan Zhang, Weiqiang Ren, Zhang Zhang, Zhen Jia, Liang Wang, and Tieniu Tan,
\newblock ``Focal and efficient iou loss for accurate bounding box
  regression,''
\newblock {\em Neurocomputing}, vol. 506, pp. 146--157, 2022.

\bibitem{siou}
Zhora Gevorgyan,
\newblock ``Siou loss: More powerful learning for bounding box regression,''
\newblock {\em arXiv preprint arXiv:2205.12740}, 2022.

\bibitem{fl}
Tsung-Yi Lin, Priya Goyal, Ross Girshick, Kaiming He, and Piotr Doll{\'a}r,
\newblock ``Focal loss for dense object detection,''
\newblock in {\em Proceedings of the IEEE international conference on computer
  vision}, 2017, pp. 2980--2988.

\bibitem{torch}
Adam Paszke, Sam Gross, Soumith Chintala, Gregory Chanan, Edward Yang, Zachary
  DeVito, Zeming Lin, Alban Desmaison, Luca Antiga, and Adam Lerer,
\newblock ``Automatic differentiation in pytorch,''
\newblock 2017.

\bibitem{coco}
Tsung-Yi Lin, Michael Maire, Serge Belongie, James Hays, Pietro Perona, Deva
  Ramanan, Piotr Doll{\'a}r, and C~Lawrence Zitnick,
\newblock ``Microsoft coco: Common objects in context,''
\newblock in {\em European conference on computer vision}. Springer, 2014, pp.
  740--755.

\end{thebibliography}
\end{document}